# ACOUSTIC-TO-WORD MODEL WITHOUT OOV

Jinyu Li, Guoli Ye, Rui Zhao, Jasha Droppo and Yifan Gong

Microsoft AI and Research, One Microsoft Way, Redmond, WA 98052
{jinyli, guoye, ruzhao, jdroppo, ygong}@microsoft.com

## ABSTRACT

Recently, the acoustic-to-word model based on the Connectionist Temporal Classification (CTC) criterion was shown as a natural end-to-end model directly targeting words as output units. However, this type of word-based CTC model suffers from the out-of-vocabulary (OOV) issue as it can only model limited number of words in the output layer and maps all the remaining words into an OOV output node. Therefore, such word-based CTC model can only recognize the frequent words modeled by the network output nodes. It also cannot easily handle the hot-words which emerge after the model is trained. In this study, we improve the acoustic-to-word model with a hybrid CTC model which can predict both words and characters at the same time. With a shared-hidden-layer structure and modular design, the alignments of words generated from the word-based CTC and the character-based CTC are synchronized. Whenever the acoustic-to-word model emits an OOV token, we back off that OOV segment to the word output generated from the character-based CTC, hence solving the OOV or hot-words issue. Evaluated on a Microsoft Cortana voice assistant task, the proposed model can reduce the errors introduced by the OOV output token in the acoustic-to-word model by 30%.

## 1. INTRODUCTION

Recently, significant progress has been made in automatic speech recognition (ASR) when switching the acoustic model from deep neural networks (DNNs) to long short-term memory (LSTM) [1][2] recurrent neural networks (RNNs) which can better model the speech sequence [3][4][5][6][7][8][9][10]. Like DNNs, LSTM-RNNs are usually trained with the cross entropy (CE) criterion, and then may be further optimized with the sequence discriminative training criterion [11][12][13][14]. Note that ASR is a sequence-to-sequence task, which maps the input speech waveform to a final word sequence or an intermediate phoneme sequence. What the acoustic modeling cares is the output of word or phoneme sequence, instead of the frame-by-frame labeling which the traditional CE training criterion optimizes. Hence, the Connectionist Temporal Classification (CTC) approach [15][16][17][18] was introduced to map the speech input frames into an output label sequence. The building network is still a LSTM-RNN, but the training objective function is changed from CE to CTC.

The most attractive characteristics of CTC is that it provides a path to end-to-end optimization of acoustic models. In the deep speech [19][20] and EESEN [21][22] work, the end-to-end speech recognition system was explored to directly predict characters instead of phonemes, hence removing the need of using lexicons and decision trees which are the building blocks in [17][18]. This is one step toward removing expert knowledge when building an ASR system. As the goal of ASR is to generate a word sequence from the speech acoustic, word unit is the most natural output unit for network modeling. In [17], the CTC with up to 27 thousand (k) word output targets was explored but the ASR accuracy is not very good, partially due to the high out-of-vocabulary (OOV) rate when using only around 3k hours training data. In [23], it was shown that by using 100k words as the output targets and by training the model with 125k hours of data, the word-based CTC, a.k.a. acoustic-to-word CTC, can beat the CTC system with phoneme unit. In [24], the training strategy of word-based CTC was explored with better initialization. The ASR task of CTC with word-based output is very simple: the output word sequence is constructed by taking the words with the maximum posterior spikes in the sequence and reducing repeated words into one if there is no blank between them. No language model or complex decoding process is involved. Therefore, the word-based CTC is a very good end-to-end ASR model. In addition to CTC, attention-based models [25][26][27] and RNN-transducers [28][29][30] are also end-to-end ASR models. Their effectiveness has been demonstrated when working with character output units. To our best knowledge, there is no report of using word output units in attention-based models and RNN-transducers. In this study, we focus on how to improve the CTC with word output units.

There are two challenges to the word-based CTC. The first one is the OOV issue. In [17][23][24], only frequent words in the training set are used as the targets and the remaining words are just tagged as the OOV. All these OOV words cannot be modeled by the LSTM-RNN and cannot be recognized during evaluation. The second issue of the word-based CTC is that it cannot handle hot-words which emerge and become popular after the network has been built. It is impossible to get satisfactory performance by directly adding output nodes in the network with the specified hot-words without retraining the network.

Inspired by the open vocabulary neural machine translation work [31], we propose an acoustic-to-word model without OOV by first building a word-based CTC in which the output vocabulary contains the frequent words in the training set together with an OOV token which all the infrequent words are mapped to. Then we train a character-based CTC by sharing most hidden layers of the word-based CTC. During recognition, the word-based CTC generates a word sequence, and the character-based CTC is only consulted at the OOV segments. Evaluated on a Microsoft internal Cortana voice assistant task, the proposed method can reduce the errors introduced by OOV output token in the acoustic-to-word model by 30%.

Although the proposed work shares the same concept of open vocabulary neural translation in [31], our work is very different from [31] as the fundament framework in our work is CTC-based speech recognition while [31] uses attention-based framework for translation. Hence, the detailed implementations are very different in these two works. There are also many works in the traditional systems which handle OOV problem for open vocabulary ASR task. Some works [32][33] only detect OOV words and recognize their phonetic transcriptions. Some studies [34][35] go further to identify the character sequence of OOV words by first recognizing the OOV word as a phoneme sequence and then using phoneme-to-character conversion to generate the character sequence. In contrast, the auxiliary character-based CTC in this study provides a much easier way to handle the OOV issue in the word-based CTC.

The rest of the paper is organized as follows. In Section 2, we briefly introduce CTC modeling and then we present the proposed acoustic-to-word model without OOV. Experimental evaluation of the algorithm is provided in Section 3. We summarize our study and draw conclusions in Section 4.

## 2. ACOUSTIC-TO-WORD CTC WITHOUT OOV

In this section, we first briefly overview the CTC modeling technology and describe the OOV issue in the acoustic-to-word CTC. Then, we propose the hybrid CTC model that can solve the OOV issue by consulting an auxiliary character-based CTC to generate candidate words. Last, we discuss how to improve the character-based CTC for better prediction.

### 2.1 CTC modeling

The CTC criterion [15] was introduced to map the speech input frames into an output label sequence [16][17][18]. To deal with the issue that the number of output labels is smaller than that of input speech frames, CTC introduces a special blank label and allows the repetition of labels to map the label sequence into a CTC path, which forces the output and input sequences to have the same length.

Denote $x$ as the speech input sequence, $l$ as the original label sequence, and $B^{-1}(l)$ represents all the CTC paths mapped from $l$. The CTC loss function is defined as the sum of negative log probabilities of all the CTC paths mapped from the correct label sequence as
$$L_{CTC} = -\ln P(l|x)$$
where
$$P(l|x) = \sum_{z \in B^{-1}(l)} P(z|x)$$
where $z$ is one CTC path. With the conditional independent assumption, $P(z|x)$ can be decomposed into a product of posterior from each frame as
$$P(z|x) = \prod_{t=1}^{T} P(z_t|x)$$
The calculation of $P(z_t|x)$ is done via the forward-backward process in [15].

The CTC output labels can be phonemes [16][17][18], characters [19][20][21][22] [36] or even words [17][23][24]. As the goal of ASR is to generate a word sequence from the speech waveform, word unit is the most natural output unit for network modeling. The recently proposed acoustic-to-word models [23][24], a.k.a. word-based CTC models, build multiple layer LSTM networks and use words as the network output units, optimized with the CTC training criterion. It is very simple to generate the word sequence with this word-based CTC model: pick the words corresponding to posterior spikes to form the output word sequence. There is neither language model nor complex decoding process involved.

However, when training the word-based CTC model, only frequent words in the training set are used as the targets and the remaining words are just tagged as the OOV. All these OOV words cannot be modeled by the network and cannot be recognized during evaluation. In next section, we proposed a hybrid CTC model to address the OOV issue and the hot-words issue discussed in the introduction.

### 2.2 Acoustic-to-word CTC without OOV

The proposed acoustic-to-word CTC without OOV model is a hybrid model which uses a word-based CTC as the primary model and a character-based CTC as the auxiliary model. The word-based CTC model emits a word sequence, and the output of the character-based CTC is only consulted at the segment where the word-based CTC emits an OOV token.

Figure 1 gives an example of the hybrid CTC model. The hybrid model has four shared hidden LSTM layers, on top of which the word-based CTC and the character-based CTC have individual one hidden LSTM layer and one softmax layer. The word-based CTC generates a sequence "play artist OOV" while the word sequence from the character-based CTC is "play artist ratatat". "ratatat" from the character-based CTC is the segment overlapped with the OOV token most, and is then used to replace the OOV token to form the final ASR output of the hybrid CTC as "play artist ratatat".

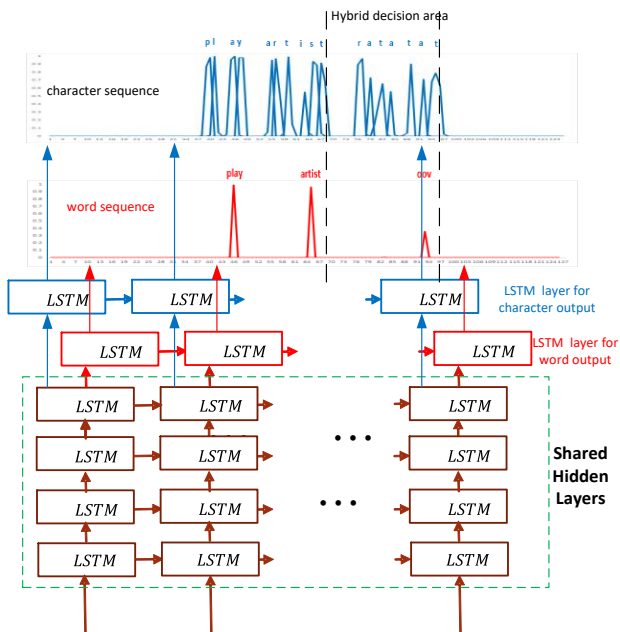

Figure 1: An example of how the hybrid CTC solves the OOV issue of the acoustic-to-word CTC.

The detailed steps for building the hybrid CTC model are described as follows
1. Build a multi-layer LSTM-CTC model with words as its output units. Map all the words occurring less than $N$ times in the training data as the OOV token. The output units in this LSTM-CTC model are all the words occurring at least $N$ times in the training data, together with OOV, blank, and silence tokens.
2. Freeze the bottom $L-1$ hidden layers of the word-CTC, add one LSTM hidden layer and one softmax layer to build a new LSTM-CTC model with characters as its output units.
3. During testing
   a. Generate the word output sequence by taking the words corresponding to maximum posterior spikes and reducing repeated words into one if there is no blank between them.
   b. If the output word sequence in 3.a doesn't contain any OOV token, then use that word sequence as the ASR result. Otherwise, proceed to the next step.
   c. Generate another word output sequence from the character-based CTC.
   d. The final ASR result is obtained by replacing the OOV token generated from the word-based CTC with the word generated from the character-based CTC that has the largest time overlap with the OOV token.

There are two ways to generate the word output sequence from the character-based CTC in step 3.c. The first way is to directly take the characters with maximum posteriors and collapse them into words. We refer this as the max output decoding. However, the character-based CTC without any decoding constraint usually gives very high WER as shown in [29]. The second way is to constrain the character-based CTC to generate only valid words (e.g., only the words in training set) using a character graph as in [36]. In this way, we avoid the character-based CTC generating invalid words. Furthermore, since the character graph includes all the words in training data, the rare words that are mapped into OOV in word-based CTC can be recognized by the character-based CTC. The character graph also allows us to handle the hot-words that emerge after the model is trained by adding the hot-words into the valid words list and reconstructing the character graph.

To measure the overlap in step 3.d, we need to define what is the segment corresponding to an output token. As blank dominates most of the time in CTC, it is not suitable to use only the frames corresponding to the token spike as the segment, which will be very short. Instead, we treat the spike frames as well as all the immediate *preceding* blank frames as the segment of the token. To get the segment corresponding to an OOV token in word-based CTC is very straightforward from the above definition. To get the segment corresponding to a word in character-based CTC, we need to first get the segment of each character with the above definition, and then concatenate all the character segments to form the word segment.

This hybrid CTC model is guaranteed to improve the accuracy of the word-based CTC because it only replaces the OOV tokens generated from the word-based CTC without changing any other word outputs. With the shared-hidden-layer structure, the alignments of words from the word-based CTC and the character-based CTC are well synchronized. Because the character-based CTC inside the hybrid CTC can generate any word without revisiting the model training, the hot-words issue can also be solved.

**2.3 Improve Character-based CTC**

The baseline character-based CTC has 28 outputs: 'a'-'z', space, blank. We refer it as the "28-character set". We need to generate word sequences from the output of character-based CTC. A word is generated by first reducing repeated characters into one and then combining all the characters except blank between two spaces. The word cannot be right if any character gets wrong. Using the context information should make the prediction of the characters better. Therefore, we add a row-convolution layer on top of the last LSTM layer as

$$\widehat{h}_t = \sum_{c=-C}^{C} W_c h_{t+c}$$

where $h_t$ is the activation vector of the last hidden LSTM layer, $W_c$ is the row convolution matrix associate with the $c$-th context hidden vector, and $2C+1$ is the total number of context hidden vectors. $\widehat{h}_t$ is then connected to the last softmax layer to predict characters. Different from the row convolution layer in [20] which only uses future hidden vectors, we use both history and future (left and right) hidden vectors to introduce more context information.

Following [36], we also construct a new character set by adding additional characters on top of the 28-character set. These additional characters include capital letters used in the word-initial position, double-letter units representing repeated characters like *ll*, apostrophes followed by letters such as *'d*, *'re* etc. Please refer to [36] for more details. Altogether such a large unit inventory has 83 characters, and we refer it as the "83-character set".

## 3. EXPERIMENTS

In this section, we use a Microsoft Cortana voice assistant task to evaluate the proposed method. The training data consists of 3400 hours of transcribed US-English Cortana audio. The test set consists of 3 hours of data from the same Cortana task. The audio data is 16k HZ sampled, recorded in mobile environments. All experiments were conducted using the computational network toolkit (CNTK) [37], which allows us to build and evaluate various network structures efficiently without deriving and implementing complicated training algorithms.

We first built a LSTM model trained with the CE criterion. The input consists of 80-dimensional log Mel-filter-bank features. It has 5 LSTM hidden layers: each has 1024 memory units and the output size of each LSTM layer is reduced to 512 using a linear projection layer [5]. There is no frame stacking, and the output HMM state label is delayed by 5 frames as in [5]. There are totally 5980 tied HMM states. This model is denoted as LSTM-CE in Table 1, with 10.05% word error rate (WER). Because of the latency restriction, we always use uni-directional models in our work.

Then, we built a phoneme-based LSTM model trained with the CTC criterion, modeling around 6000 tied context-dependent phonemes. It has the same 5-layer LSTM structure with projection layer as the previous LSTM-CE model. Eight frames of 80-dim log Mel-filter-bank features are stacked together as the input, and the time step shift is three frames as in [17]. Without mentioning explicitly, all the CTC models in this study use the same structure as this model. This model is denoted as LSTM-CTC (phoneme) in Table 1, with 9.87 % WER. Both the LSTM-CE model and LSTM-CTC (phoneme) model use a strong 5-gram language model (LM) for decoding. The gap between the LSTM-CE model and the LSTM-CTC (phoneme) model is not large, consistent with the recent report [38].

Next, we built an acoustic-to-word CTC model by modeling around 27k most frequent words in the training data. These frequent words occurred at least 10 times in the training data. All other infrequent words are mapped to an OOV output token. This model, LSTM-CTC (word), gets 13.59% WER, among which the OOV token contributes 1.70% WER. In other words, if every OOV token can be converted to the right word, the WER will be reduced to 11.89%. Note that the WER gap between the phoneme-based CTC and the word-based CTC is not small because the word-based CTC doesn't use any LM while the phoneme-based CTC uses a very strong LM trained from much larger amount of text than the 3400hr speech transcription. The WER gap is consistent with what has been observed in [17][24]. All the CTC models except the phoneme-based CTC model in this study purely rely on the network score to generate outputs without using LM.

**Table 1:** WER comparison of baseline LSTM-CE, LSTM-CTC (phoneme), and LSTM-CTC (word)

| Model | WER (%) |
|---|---|
| LSTM-CE | 10.05 |
| LSTM-CTC (phoneme) | 9.87 |
| LSTM-CTC (word) | 13.59 |

We use the structure in Figure 1 to build hybrid CTC models. The first step is to build character-based CTC models by sharing 4 hidden LSTM layers of the word-based CTC model. On top of the shared hidden layers, we add a new LSTM hidden layer and a softmax layer to model character outputs. The output units of the character-based CTC can be from either the 28-character set or the 83-character set described in Section 2. When training the character-based CTC model, only the added LSTM hidden layer and softmax layer are updated. The bottom 4 hidden LSTM layers are not updated because they are shared with the word-based CTC. Next, the character-based CTC model is improved with row convolution described in Section 2.3. The row convolution operates on 9 frame hidden vectors from the last LSTM layer, with 4 history frames, a central frame, and 4 future frames.

**Table 2:** WER comparison of character-based CTC models

| Model | WER (%) |
|---|---|
| CTC (28-character, max output) | 33.79 |
| CTC (28-character) | 23.87 |
| CTC (28-character + row convolution) | 20.83 |
| CTC (83-character) | 20.25 |
| CTC (83-character + row convolution) | 18.91 |

Table 2 gives the WER of different character-based CTC models. The baseline CTC with the 28-character set has 33.79% WER when just using the max output decoding which picks the characters with maximum posteriors and then collapses to words. Such a high WER is consistent with what has been observed in other sites [29]. Adding the constraint that only valid words from training set can be generated, the WER is reduced by 10% absolute to 23.87% WER. In the following, the default decoding setup of the character-based CTC is with character-graph decoding with the valid word list constraint. Clearly, the vanilla character-based CTC is far behind the word-based CTC, and hence can only be used as an auxiliary model. By taking 9-frame hidden vector context with row convolution, the character-based CTC can be improved to 20.83% WER. Then, the CTC with the 83-

character set improves its counterpart with the 28-character set from 23.87% WER to 20.25% WER. Finally, the CTC model with the 83-character set and row convolution gets 18.91% WER, still 5% absolute higher than the WER from the word-based CTC.

The row convolution method can get 12.74% relative WER reduction (from 23.87% WER to 20.83% WER) with the 28-character set, but only gets 6.62% relative WER reduction (from 20.25% WER to 18.91% WER) with the 83-character set. One reason is that the 83-character set also somehow handles the context information (e.g., with double letters), which is also handled by the row convolution method.

Table 3 gives several examples showing how the row convolution method helps to improve the WER of the CTC with the 28-character set. Without consulting context frames, the CTC model sometimes misses several characters while the row convolution model can emit the right words out based on its context.

**Table 3:** Examples that the CTC with row convolution gets the right recognition result.

| CTC (28-character) | CTC (28-character + row convolution) |
|---|---|
| how much **one** | how much **money** |
| wake me up in **a** hour | wake me up in **an** hour |
| tell me good joke | tell me **a** good joke |

Table 4 shows how the CTC with the 83-character set is better than the CTC with the 28-character set with several examples. Modeling the double letters helps to win these examples.

**Table 4:** Examples that the CTC with the 83-character set gets the right recognition result.

| CTC (28-character + row convolution) | CTC (83-character + row convolution) |
|---|---|
| my wife is **ten my** | my wife is **tammy** |
| okay **jail** | okay **jill** |
| **kellogs** | **kellogg's** |

Table 5 gives the WERs of hybrid CTC models. When the CTC with 28 characters is used with max output decoding, the hybrid CTC only slightly improves the baseline word-based CTC because such a character-based CTC setup cannot give too much helps due to high WER as shown in Table 2. When decoding with character graph constrained by valid words, the hybrid CTC obtains around 13.09% WER, with 0.5% absolute WER reduction from the word-based CTC. Because the OOV token brings 1.7% absolute WER to the word-based CTC model, this means the hybrid CTC can recover 30% errors introduced by the OOV token. It is somehow surprising that although both the row convolution CTC modeling and the CTC with the 83-character set have better WER than the CTC with the 28-character set, neither setup can help the final WER of the hybrid CTC.

**Table 5:** WER comparison of hybrid CTC models

| Model | WER (%) |
|---|---|
| CTC (word) | 13.59 |
| CTC (word) + CTC (28-character, max output) | 13.42 |
| CTC (word) + CTC (28-character) | 13.09 |
| CTC (word) + CTC (28-character + row convolution) | 13.10 |
| CTC (word) + CTC (83-character + row convolution) | 13.08 |

In Table 6, we show how the hybrid CTC model performs with some examples. The first three are the examples that the hybrid CTC can recover the right words from the OOV token. "azusa", "ratatat", and "wanna", all these addresses and names, are the words not in the frequent words in the training set, and haven't been modeled by any output node in the word-based CTC model. However, they can be successfully recovered by the character-based CTC.

The last three rows in Table 6 are the examples that the hybrid CTC still fails to recover the right words from the OOV token. "margera" is recognized as "marger" by the character-based CTC. Such error happens with one character missing, revealing the weakness of character-based CTC. "purr" is recognized as "per", and "kristi" is recognized as "christi" by the character-based CTC. These errors are homophone errors, which cannot be handled by character-based CTC unless high level information is blended into the decision.

**Table 6:** Examples of the outputs of word-based CTC and hybrid CTC models (CTC (word) + CTC (28-character))

| Reference | Word-based CTC | Hybrid CTC |
|---|---|---|
| costco **azusa** | costco OOV | costco **azusa** |
| play artist **ratatat** | play artist OOV | play artist **ratatat** |
| text mara **wanna** | text mara OOV | text mara **wanna** |
| april **margera** | april oov | april **marger** |
| why does my kitty **purr** | why does my kitty OOV | why does my kitty **per** |
| call **kristi** matthews | call OOV matthews | call **christi** matthews |

In Table 5, it is a little disappointing that neither row convolution nor 83-character set modeling improves the final WER of the hybrid CTC. We also examined the results and found that most of times these two methods help to improve the recognition results that the word-based CTC succeeds. For the failed cases in Table 6, they cannot help too much. For example, "margera" is recognized as "marger" by the CTC with the 28-character set, and recognized as "marera" by the CTC with the 83-character set and row convolution. They also cannot help the homophone error cases. Even with better modeling, it is sometimes still very challenging for the character-based CTC to get words right for the cases that the word-based CTC fails.

## 4. CONCLUSIONS AND FUTURE WORKS

In this paper, we have presented a hybrid CTC model that solves the OOV issue and the hot-words issue presented in the acoustic-to-word CTC models, a.k.a. the word-based CTC, by using the output from the word-based CTC as the primary ASR result and consulting the character-based CTC at the segment where the word-based CTC emits an OOV token. By only replacing the OOV tokens with the words generated from the character-based CTC, the proposed method is guaranteed to improve the accuracy of the acoustic-to-word CTC. The shared hidden layer structure helps to align the word segments between the word-based CTC and the character-based CTC so that the OOV token lookup algorithm can work. Evaluated on a Microsoft Cortana voice assistant task in which the word-based CTC has 1.7% WER introduced by the OOV token, the hybrid CTC model can reduce 0.5% absolute WER, representing a recovery of 30% errors caused by the OOV token.

Several research issues will be addressed in the future to further increase the effectiveness of the algorithm presented in this paper. First, a better character unit set should be considered to improve the accuracy of the character-based CTC model. Recently, gram-CTC [39] was proposed to automatically learn the most suitable decomposition of target sequences, which not only boosts the modeling flexibility but also improves the final ASR accuracy. We are now trying to incorporate the gram-CTC into our system. Second, the character-based CTC has very high WER (around 33%) when using the maximum output decoding. We add valid word constraint when generating words from the character-based CTC and bring down its WER to 23% so that the words used to replace OOV tokens are useful. However, a character-based CTC model with decoding constraint is not a clean end-to-end model as it still involves expert knowledge. We are now pursuing more advanced method which can improve the character-based WER to as low as 18% with the maximum output decoding [40]. Last, with thousand hours of training data, the word-based CTC still has an accuracy gap from the phoneme-based CTC, which has been observed from various sites. We found that the word-based CTC can significantly improve the accuracy of the phoneme-based CTC by combining them together, given very different error patterns from these CTC models. Therefore, it is meaningful to invest on the word-based CTC even from the production point of view. At the same time, we are working on improving the modeling of word-based CTC so that we can deploy such an end-to-end acoustic-to-word model to production.